\documentclass[
]{ceurart}

\usepackage{listings}
\begin{document}

\copyrightyear{2023}
\copyrightclause{Copyright for this paper by its authors.
  Use permitted under Creative Commons License Attribution 4.0
  International (CC BY 4.0).}

\conference{Non official and non reviewed paper for CheckThat'23}

\title{KUCST at CheckThat 2023:\\How good can we be with a generic model?}


\author{Manex Agirrezabal}[%
orcid=0000-0001-5909-2745,
email=manex.aguirrezabal@hum.ku.dk,
url=https://manexagirrezabal.github.io/,
]
\address{Centre for Language Technology (CST), Department of Nordic Studies and Linguistics, University of Copenhagen, Emil Holms kanal 2, 2300 Copenhagen, Denmark}

\begin{abstract}
In this paper\footnote{The author of the paper could not submit the paper on time for various reasons, but the system did appear in the official CheckThat rankings.} we present our method for tasks 2 and 3A at the CheckThat2023 shared task. We make use of a generic approach that has been used to tackle a diverse set of tasks, inspired by authorship attribution and profiling. We train a number of Machine Learning models and our results show that Gradient Boosting performs the best for both tasks. Based on the official ranking provided by the shared task organizers, our model shows an average performance compared to other teams.
\end{abstract}

\begin{keywords}
  LaTeX class \sep
  paper template \sep
  paper formatting \sep
  CEUR-WS
\end{keywords}

\maketitle

\section{Introduction}
Checkworthiness, Subjectivity, Political Bias, Factuality, and Authority. Those are interest areas of the CheckThat! Lab, which is the responsible of organizing a series of competitions where different research groups are asked to work on a challenge. In this work we decided to put our efforts in two of those areas: subjectivity and political bias.

In the era of digital media and news, it is very important to ensure the access to objective news. Failure to comply with this, may compromise democracy, as subjective information can be used to shape the opinion of their recipients (readers, listeners, and so on). Objectivity is, thus, a requirement for a democratic society. This work is an attempt to provide a model that can detect whether a sentence is objective or subjective.

If we move the discussion topic to a political side, political biases are still a concern in society, especially now that broadcasting any type of content is easier than previously. Let us suppose for a moment that we are able to make sure that all news are objective. Even though we make sure that all news are objective, this may still not mitigate political bias.
It is still possible to have politically biased articles if, for instance, some information is missing, or there is no fact-checking. It is, therefore, very important to control for these issues for a more democratic society.



In this paper we present our approach for tasks 2 \cite{clef-checkthat:2023:task2} and 3A \cite{clef-checkthat:2023:task3} at the CheckThat2023 shared task \cite{checkthat}. Our approach is based on features inspired by the authorship attribution and profiling community.

This paper is structured as follows. First we review some works on subjectivity detection and political bias detection. Later we present the data used and methods developed. After that we show the results of different models and we further discuss them. We then conclude the paper and propose some possible future directions.

\section{Related work}
\subsection*{Subjectivity detection}
Das and Bandyopadhyay \cite{das2010subjectivity} proposed a subjectivity detection model that incorporates a combination of lexico-semantic, syntactic, and discourse-level features. Their approach includes features such as POS tags, Sentiwordnet, word stems, syntactic information, chunks, dependency information, and discourse-level features like the title of the text, first paragraph, and last two sentences. Additionally, they use genetic algorithms for optimization. Notably, their experimental results demonstrate high precision and recall rates exceeding 90\% for the English language, with a particularly high recall. Because of this, this model stands out as one of the most performant approaches to date, offering valuable insights for future research directions. In our work, however, we have focused on prioritizing genericness, simplicity, and interpretability.

In the field of subjectivity detection, Lin and He \cite{lin2011sentence} introduced \textit{subjLDA}, a model that leverages lexical clues and weak supervision. Initial findings indicate that subjLDA outperformed the widely used LDA model in terms of various evaluation metrics, with the exception of slightly lower objective recall. Furthermore, subjLDA demonstrated comparable performance to a previously proposed bootstrapping approach, despite utilizing a significantly smaller training set. These results are highly intriguing, particularly given the specific requirements of our task. SubjLDA would have been a reliable and suitable choice, but as our interest is to check the performance of our generic model, we will not make use of it.

Following recent trends, in the work by \cite{huo2020utilizing} they employed BERT and different fine-tuning strategies for the detection of subjectivity in the data by \cite{pang2004sentimental} and \cite{hube2019neural}

In their work, Sagnika \cite{sagnika2021attention} and colleagues tested the efficacy of CNN-LSTM networks with attention for subjectivity detection. They represent sentences as word embeddings that include sentiment and part-of-speech information. They show that incorporating attention in their proposed model improves its peformance, which is because the model can remember long range relationships and dependencies within the sentences \cite{sagnika2021attention}.

\subsection*{Political bias detection}
Baly and colleagues \cite{baly2020we} conducted notable work in political bias detection by constructing a well-balanced dataset encompassing a diverse range of topics and media sources. They ensured the effectiveness of their models by testing them with examples from unseen media during training. This approach prevents the models from merely learning to detect the source of the target news article, instead emphasizing the prediction of its political ideology. The authors proposed several sophisticated models, including ones based on adversarial media adaptation and a unique triplet loss framework to discourage the models from predominantly modeling the source information rather than the political bias present in the news article.

Chen et al. \cite{chen2020analyzing} conducted an analysis of political biases in news articles using Recurrent Neural Networks with Gated Recurrent Units (GRU). In their study, they employed a reverse feature analysis to gain further insights into the results. To interpret the findings, they utilized various tools, including LIWC (Linguistic Inquiry and Word Count) features \cite{pennebaker2015development}. Their investigation revealed an interesting observation that the last quarter of news articles appeared to be the most biased section. This finding sheds light on the potential significance of the concluding parts of news articles in terms of political bias.

There have been works in the development of political bias detection models for other languages. For instance Gangula and colleagues \cite{gangula2019detecting} we created a dataset that consisted of 1329 news articles from various Telugu newspapers. Their task, which involves classifying bias towards political parties rather than the typical left, right, or center categorization, introduces an additional layer of complexity. This approach may prove beneficial as different political parties often exhibit distinct styles, and their model may capture these nuances. The authors utilized a headline encoder, an article encoder, and an attention layer. Subsequently, they applied softmax to make predictions regarding the associated political party. Comparing this architecture to several less complex baseline methods, their final model incorporating attention outperformed all the baselines.

In the work by Aksenov and colleagues, \cite{aksenov2021fine} a model for political bias detection specifically designed for German language. They constructed a dedicated dataset for this task and conducted experiments using various features, including Bag-of-Words, TF-IDF, and BERT. Four different classification models were trained: Logistic Regression, Naive Bayes, Random Forest, and EasyEnsemble. Notably, they conducted an additional analysis to identify the most relevant words associated with different bias classes. This work closely aligns with our own approach, making it highly relevant to our research.

\section{Data and Mathod}
We utilized the dataset provided by the organizers of the shared task \cite{checkthat,clef-checkthat:2023:task2,clef-checkthat:2023:task3}. The relevant details and statistics are presented in the table below. Task 2 consists of approximately 1,000 instances, wherein sentences from news articles are annotated as either objective or subjective. Notably, around 60\% of the sentences are labeled as objective, while the remaining 40\% are classified as subjective. In Task 3A, we have access to a larger corpus of approximately 45,000 instances, comprising news articles and their corresponding headlines. Each article in Task 3A is annotated with a political bias, indicating whether it leans towards the center, left, or right ideology.

    \begin{tabular}{l|r|r|r|r}
    Task & \#classes & \#instances (train)& \#instances (test) & FE. time (mins)\\
Task 2 & 2 & 702 & 347 & 1.4 \\ 
Task 3A & 3 & 30197 & 14874 & 94.35\\ 
\end{tabular}

In our approach to address the task at hand, we followed a general methodology that we anticipated would be applicable to various tasks, including bot vs. human discrimination, subjectivity detection, sexism detection, and others. While our original intention was to compare feature importance across different tasks, this article primarily emphasizes the characteristics of subjectivity and political bias detection.

We employed a range of commonly used stylometric features, drawing inspiration from the works of Koppel and Schler \cite{koppel2009computational} and Stamatatos \cite{stamatatos2009survey}. Specifically, our feature set included the following:

\begin{itemize}
\item Word-level Bag-of-Words (unigrams): This feature representation captures the frequency of individual words in the input text.
\item TF-IDF Weighted Bag-of-Words (unigrams): This feature representation applies TF-IDF weighting to the word-level Bag-of-Words, considering both the term frequency and its inverse document frequency.
\item Character-level Bag-of-Words (1-4grams): This feature representation captures the presence and frequency of character-level n-grams, ranging from 1 to 4 characters in length.
\item POS Tag Bag-of-Words (1-4grams): This feature representation encodes the part-of-speech tags of the text, considering both the presence and frequency of different POS n-grams.
\item Morphological Features: We utilized the Stanza package, specifying the appropriate model, to extract morphological features from the input text.
\item BERT Encoding: To capture deeper contextual information, we employed the BERT-based model \cite{devlin-etal-2019-bert}, specifically the \texttt{bert-base-cased} model, to generate embeddings for the input text.
\end{itemize}

We train seven different classifiers: K-Nearest Neighbors ($K=5$), Logistic Regression, Linear Support Vector Machine, Multilayer Perceptron, Decision Tree, Random Forest and Gradient Boosting. All models are trained using a Train/test validation procedure, where two thirds of the data are used for training and the remaining part is used for testing (0.66 vs 0.33).

\section{Results and discussion}
Please find in Table \ref{tab:checkthat2} and \ref{tab:checkthat3a} the performance of different classifiers in the two tasks, together with a majority baseline. We can observe that all models surpass the majority baseline in terms of the weighted F1-score. Gradient Boosting is the best classifier in both tasks and KNN seems to perform the poorest in both cases as well.

We find relevant to briefly mention the training time, as this might be a reason to select or avoid certain classifiers. In the case of Task 2, as the amount of data is relatively small, all classifiers were trained in less than one minute. However, in task 3A, as there were ~30,000 instances for training, it took around one hour for training the Gradient Boosting classifier. All other methods required less than 5 minutes. If training time was to be a relevant matter when deciding the best classifier, the next possible choice could be the Logistic Regression or the Multilayer Perceptron.

\begin{table}[]
\begin{tabular}{|l|rrrrrrrr|}
\hline
\textbf{}                 & \textbf{Majority} & \textbf{KNN} & \textbf{LR} & \textbf{LSVM} & \textbf{MLP} & \textbf{DT} & \textbf{RF} & \textbf{GB}        \\
\hline
\textbf{weightedf1s}      & 0.459866       & 0.61393      & 0.708894    & 0.704949      & 0.682075     & 0.640001    & 0.719856    & \textbf{0.769598} \\
\textbf{accuracy}         & 0.608069       & 0.622478     & 0.714697    & 0.711816      & 0.688761     & 0.639769    & 0.729107    & \textbf{0.772334} \\
\textbf{Training time} & $<1s$ & $<1s$ & $<1s$ & $<1s$ & $<30s$ & $<1s$ & $<1s$ & $<30s$ \\
\hline
\end{tabular}
\caption{Check that task 2 results (2 labels: OBJ, SUBJ)}
\label{tab:checkthat2}
\end{table}

\begin{table}[]
\begin{tabular}{|l|rrrrrrrr|}
\hline
\textbf{}                 & \textbf{Majority} & \textbf{KNN} & \textbf{LR} & \textbf{LSVM} & \textbf{MLP} & \textbf{DT} & \textbf{RF} & \textbf{GB}        \\
\hline
\textbf{weightedf1s}      & 0.218145       & 0.419038     & 0.521688    & 0.455131      & 0.524287     & 0.453776    & 0.502435    & \textbf{0.559401} \\
\textbf{accuracy}         & 0.38927        & 0.421541     & 0.523262    & 0.510354      & 0.522791     & 0.451459    & 0.5119      & \textbf{0.561382} \\
\textbf{Training time} & $<1s$ & $<1s$ & $<30s$ & $<5m$ & $<5m$ & $<1m$ & $<1m$ & $<1h$ \\
\hline
\end{tabular}
\caption{Check that task 3A results (3 labels: left, center, right)}
\label{tab:checkthat3a}
\end{table}

Based on official results from the shared task organizers, in Task 2, our model ranked 4th out of 11 teams with a Macro F1-score of $0.73$. This indicates a reasonably good performance on the given task. The results for Task 3A reveal a rather high Mean Absolute Error (MAE) of 0.736. Our model secured the 3rd position out of 5 participants on the leaderboard. These results highlight the challenging nature of the task.

In Task 2, we observed that TFIDF features consistently yielded poor results. This finding suggests that the topic of a text may not significantly influence the likelihood of it being subjective or objective. However, further investigations are necessary to validate this hypothesis conclusively.

Additionally, we noticed that character Bag-of-Words (BOWs) features exhibited relatively strong performance in both Task 2 and Task 3A. However, the exact reasons behind their effectiveness remain challenging to interpret.

Our study utilized simple features, primarily focusing on authorship attribution and profiling features. It is important to note that these features may not be fully representative or capture the entirety of the complex dynamics associated with the subjectivity and objectivity of texts. Future research endeavors should consider incorporating more comprehensive and diverse feature sets to enhance the accuracy and robustness of the models.

\section{Conclusion and Future Work}
In this work we present our approach for the detection of subjectivity and political bias. We employed a generic model inspired by authorship attribution and profiling, and tested with a number of classifiers. Our results show that the Gradient Boosting method is the best performing one. Our model ranked 4th out of 11 teams in task 2, and 3rd out of 5 teams in task 3A.

As our model was not specifically designed to perform subjectivity detection or political bias detection, there are many improvements that can be done for this specific task by including domain-specific features. For instance, as the last portion of news articles tends to be more subjective \cite{chen2020analyzing}, this information could be exploited.

Besides, we believe it would be interesting to make an attempt to predict what is the subjective portion of a news article when this is labeled as subjective. This could be done in a similar way that structural sentiment analysis was performed \cite{barnes-etal-2022-semeval}.

To conclude, we believe that the approach followed by Lin and He \cite{lin2011sentence}, where they make use of lexical clues for subjectivity detection is a good future direction, especially because the limited size of our corpus.

\bibliography{sample-ceur}

\end{document}